\documentclass[twoside,leqno,twocolumn]{article}

\usepackage[letterpaper]{geometry}
\usepackage{cite}
\usepackage{amsmath,amssymb,amsfonts}
\usepackage{algorithmic}
\usepackage{graphicx}
\usepackage{tabularx}
\usepackage{textcomp}
\usepackage{multirow}
\usepackage{xcolor}
\usepackage{caption,subcaption, float}
\usepackage{bm}
\usepackage{enumitem}
\usepackage{ltexpprt}

\usepackage{hyperref}

\begin{document}


\title{\Large Semi-supervised Classification using Attention-based Regularization on Coarse-resolution Data}

\author{
Guruprasad Nayak\thanks{These authors contributed equally to this article}  ${^\dagger}$, 
Rahul Ghosh$^*$${^\dagger}$, 
Xiaowei Jia\thanks{University of Minnesota. \{nayak013,ghosh128,jiaxx221, kumar001\}@umn.edu}, 
Varun Mithal\thanks{LinkedIn varunmithal@gmail.com},  
Vipin Kumar ${^\dagger}$
}
\date{}
\maketitle

\begin{abstract}
    Many real-world phenomena are observed at multiple resolutions. Predictive models designed to predict these phenomena typically consider different resolutions separately. This approach might be limiting in applications where predictions are desired at fine resolutions but available training data is scarce. In this paper, we propose classification algorithms that leverage supervision from coarser resolutions to help train models on finer resolutions. The different resolutions are modeled as different views of the data in a multi-view framework that exploits the complementarity of features across different views to improve models on both views. Unlike traditional multi-view learning problems, the key challenge in our case is that there is no one-to-one correspondence between instances across different views in our case, which requires explicit modeling of the correspondence of instances across resolutions. We propose to use the features of instances at different resolutions to learn the correspondence between instances across resolutions using an attention mechanism.Experiments on the real-world application of mapping urban areas using satellite observations and sentiment classification on text data show the effectiveness of the proposed methods.
\end{abstract}

\section{Introduction}







\begin{figure}[H]
    \centering
    \begin{subfigure}{\linewidth}
        \centering
        \includegraphics[width=\linewidth]{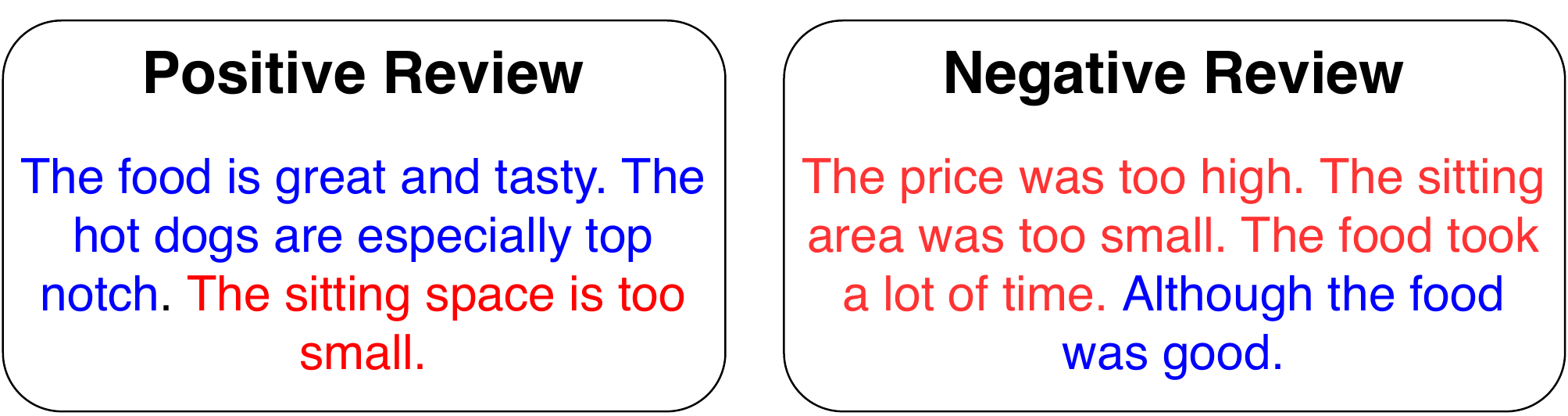}
        \caption{Natural Language Processing: Sentiment Classification. Colors blue and red denote sentences with positive and negative sentiment respectively.}
        \label{fig:Multires_NLP}
    \end{subfigure}%
    
    \begin{subfigure}{\linewidth}
        \centering
        \includegraphics[width=\linewidth]{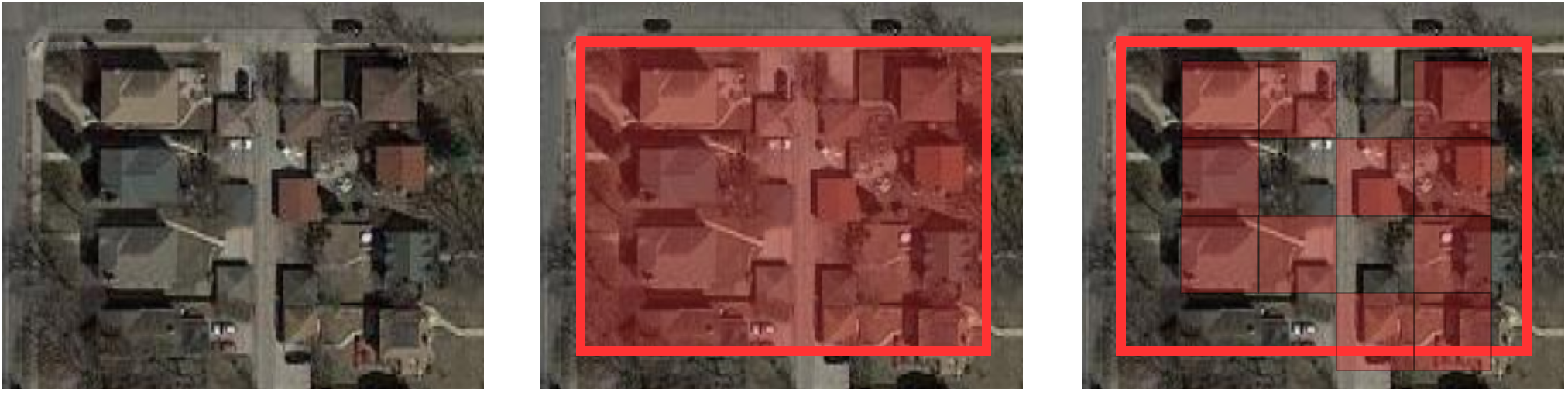}
        \caption{Remote sensing: Urban mapping. The middle image shows the extent of a coarse-resolution pixel marked as urban. The right image shows the constituting fine resolution instances that are urban.}
        \label{fig:Multires_RS}
    \end{subfigure}%
    
    \caption{Multi-resolution data sets for different application domains. In many cases, although the eventual goal is to label fine-resolution instances, obtaining sufficient supervision for fine resolution might be difficult. Coarser resolution supervision and unlabeled data might be relatively easier to obtain.}
    \label{fig:Multires_examples}
\end{figure}

In many applications, supervised learning problems can be studied at different resolutions of observation. Data sets collected at different resolutions might differ in terms of the kind of features that they have and the availability of training samples at that resolution for the particular learning problem. Figure \ref{fig:Multires_examples} shows the multi-resolution nature of data from two different example domains, namely, natural language processing and remote sensing. The first example considers the task of sentiment classification \cite{SentimentMultiRes} on restaurant reviews. Sentiments can be analysed at the sentence-level or at the paragraph-level. The second example considers a multi-resolution spatio-temporal data set\cite{AtluriSurvey} for the task of mapping urban areas using satellite-collected observations of locations on Earth. Satellites observe the Earth in a variety of spatial and temporal resolutions. Note that although a location might be urban at 30m$\times$30m resolution, the constituting 10m$\times$10m resolution instances may not all be urban, as shown in figure \ref{fig:Multires_RS}. Similar examples of multi-resolution data can be seen in many other domains such as drug discovery \cite{MILdrug},  image retrieval \cite{MILir} and text categorization \cite{MILtext}.
The performance of a predictive model is strongly dependant on the size and representativeness of the samples used to train it. One of the key challenges in supervised learning is the paucity of labeled training data. This is particularly prominent when we are trying to learn a model for fine resolution predictions. For example, in the sentiment-classification task, it is relatively cheaper to obtain training labels for the entire review than for individual sentences within the review. Moreover, the number of samples required to train an accurate model increases with (a) complexity of the chosen classifier and (b) the heterogeneity of the feature space. For example, consider the example of using satellite images to map the extent of urban areas on the surface of the Earth. Urban areas (and non-urban areas) are heterogeneous enough globally that they require complex models like deep neural networks to predict them accurately. This creates the need for large number of training samples which can be hard to meet in this application since manual annotation is the primary source of obtaining labeled data. Similar challenges occur in other application areas of predictive modeling where the scale and variety of data instances is large.


In this paper, we present a novel method to enhance the classification model trained with limited supervision on fine resolution using auxiliary information from coarser resolutions.
Specifically, we assume that: 1) we have limited labeled data at both fine and each of the coarser resolutions and; 2) we have abundant unlabeled data across all resolutions.
By leveraging the abundant unlabeled data available across resolutions, we can enforce consistency between the model trained on the fine resolution with the models trained on coarser resolutions. This acts as a regularizer for the fine-resolution model thereby increasing its robustness and generalizability.
For example, in the sentiment classification case, we might have some positive and negative-sentiment sentences and similarly, some  positive and negative-sentiment reviews (paragraphs). In addition, we might have a large text corpus with features for both reviews and constituting sentences (multi-resolution unlabeled data). 

The consistency between the fine resolution and the auxiliary coarser resolutions can be enforced by framing it as a multi-view learning \cite{MVLsurvey} problem, where different resolutions can be seen as different \emph{views} that describe the same data set. However, the key challenge in applying the multi-view learning approach in our case is the absence of one-to-one correspondence between instances from one view to the other. Furthermore, the number of fine-resolution instances that take the same class label as the corresponding coarse-resolution instance can be variable and may not be determined beforehand. We propose two strategies to handle this many-to-one correspondence between a pair of resolutions, coarse and fine: (1) Using a Multiple Instance learning (MIL) \cite{MILSurvey2} approach, where the presence of even a single instance of the class of interest among the finer resolution instances makes the corresponding coarse resolution instance positive; (2) Using the features of the coarse resolution instance and the corresponding fine resolution instances to compute the matching instances through an attention mechanism \cite{AttentionNLP}. While the MIL strategy only allows the most positive fine-resolution instance to influence the corresponding coarse-resolution instance, the attention mechanism is more flexible and allows multiple fine resolution instances within a coarse-resolution instance to exert influence by assigning a relevance score to each based on their feature values.

\subsection{Contributions}
To summarize, we make the following contributions in this work:
\setlist{nolistsep}
\begin{enumerate}
    \item We formalize the use of auxiliary coarse-resolution labeled and unlabeled data to enhance fine-resolution classification models by considering it as a multi-view learning problem.
    \item We propose a Multiple Instance Learning (MIL) strategy to handle the many-to-one correspondence between instances of the fine resolution and instances of the auxiliary coarse resolutions. This method uses a stricter \emph{presence-based} assumption that considers a coarse-resolution instance positive-labeled if at least one of the constituting fine resolution instances is positive-labeled. 
    \item We further propose a superior strategy to model the many-to-one correspondence, that considers more than one fine resolution instance to influence a given coarse resolution instance. This strategy uses an attention mechanism that assigns relevance to each fine resolution instance based on its feature value and the feature value for the coarse resolution instance.
    \item Our proposed methods demonstrably improve the classification performance on fine-resolution instances for data sets from multiple applications: (1) land cover mapping, which is a problem of great environmental significance and; (2) sentiment classification which is an important problem in natural language processing
\end{enumerate}

\subsection{Code and Data repository} The code for the algorithms proposed in this paper along with the data sets used for evaluation can be found at the following link - \href{https://github.com/2021rahul/Multi-view-Regularization-using-Attention-Mechanism}{https://github.com/2021rahul/Multi-view-Regularization-using-Attention-Mechanism}

\section{Related Work}
\textbf{Semi-supervised learning} leverages abundant unlabeled data in addition to limited number of labeled training data to boost the performance of the learning algorithm. This involves some kind of assumption on the unlabeled data that is typically expressed as a regularization. A common example is the cluster assumption which states that labels should not change within clusters of unlabeled points. As another example, manifold regularization \cite{SSRManifold} assumes that unlabeled instances that are similar in a transformed low dimensional feature space should have similar predictions. Transductive SVMs \cite{TSVM1} learn decision boundaries should lie in low density regions in the feature space. 

Our work can also be viewed as an example of weakly-supervised learning \cite{zhouWLIntro,thesis,SentimentMultiRes}. In a weakly-supervised learning scenario, we have very few training samples that have exact labels corresponding to the target variable. However, we have plenty of weakly-labeled instances i.e we have an imperfect version of the target variable for these instances. The idea is that, by modeling the imperfection in the weak labels, we can mitigate the lack of (strongly-labeled) training data. However, the kind of weak supervision present in our problem setting, has not been studied previously, to the best of our knowledge.

\textbf{Attention models}, first introduced for the task of neural machine translation \cite{bahdanau2014} have become an essential part of neural network models in many applications including Natural Language Processing, Statistical Learning, Speech and Computer Vision. The central idea of attention modeling is to selectively influence the prediction on a query instance based on other key instances. This is done by computing an \emph{alignment score} between the query instance and each one of the key instances. The output score for the query instance is the sum of the values for the key instances, weighted by their respective alignment scores with the query instance. Recently, attention models have been used in classification problems with group-labels, where introducing attention mechanism has been shown to increase interpretability for the group-level prediction with respect to the underlying instances \cite{attentionMIL}. Another instance of using attention mechanism to capture hierarchy is in \cite{hierarchyAttention}, where a neural network is designed for document classification that incorporates the different hierarchical levels in the document, namely - word, sentence, document. At each level of the hierarchy, an attention mechanism selectively passes information from the most relevant instances of the lower hierarchy to the higher one. Both of these lines of work that use attention to model hierarchy work exclusively with coarse-resolution labels and focuses on coarse-resolution predictions, which is different from the problem setting in this paper, where our goal is to make predictions at the fine-resolution.

\section{Method}

\subsection{Problem Setting}
In this paper, we consider a classification problem, in which we are trying to predict a binary label at a fine resolution $R_0$, using auxiliary information from coarser resolutions $\{R_1,\cdots,R_K\}$. Data instances at every resolution $0 \leq k \leq K$ can be described through attributes $(\bm{x}, l, y)$ defined as follows
\setlist{nolistsep}
\begin{enumerate}
    \item Features $\bm{x} \in R^{D_k}$, where ${D_k}$ is the dimensionality of the features observed in the $k$th resolution
    \item Location $l$ that describes the position of the observation. For example, in spatio-temporal data sets, this encodes the location in space and the point in time where each instance is observed. This location information is used to assign correspondence between instances across resolutions.
    \item Label $y \in \{0,1\}$ that encodes the presence ($y=1$) or absence ($y =0$) of the class of interest at location $l$
\end{enumerate}
\textbf{Goal: }To learn classification model $\bm{w}_0$ at the fine resolution $R_0$ that can take the features $\bm{x}^0$ observed for an instance on that resolution and predict its corresponding label $y^0$. \\ \hfill
\textbf{Training data: }During the training phase, for each resolution $k$, we are provided a set $T_l^k$ of $N_l^k>0$ labeled samples $\{ \bm{x}^k_i, l^k_i, y^k_i\}_{i=1}^{N_l^k}$. In addition, we have a large region of unlabeled data where observations are available at every resolution i.e.; for each resolution $0 \leq k \leq K$, we have features $\bm{x}^k_i$ for every observable location $l$ within this region, forming a unlabeled training data set $T_u^k$ of $N_u^k$ samples $\{ \bm{x}^k_i, l^k_i\}_{i=1}^{N_u^k}$.

\subsection{Multi-view framework}
The key idea of this paper is to regularize the fine resolution model trained with limited supervision by enforcing consistency of its predictions on unlabeled data with the predictions of models trained on other coarser resolutions. To this end, we learn classification models $\bm{w}_k$ at every resolution $k \in \{0 \cdots, K\}$. In particular, classifier $f_{k}(\bm{x^{k}}; \bm{w}_{k})$ at resolution $k$ with parameters $\bm{w}_k$ models $Pr(y^k =1 | \bm{x}^k)$, where $y^k$ is the label at that resolution. We place no restriction on the actual form of $f$. It could take any form like LSTMs for sequence data or CNNs for spatial/image data or more traditional classifiers like a neural network with few hidden layers. However, instead of learning the classifiers at different resolutions independently, as would be the case in a conventional approach, we propose to use the large number of unlabeled data available on the same region of the field of observation to enforce consistency in predictions across resolutions and thus, make the models on different resolutions learn from each other. Thus, the objective function takes the following form,
\begin{equation}
    \label{eqObjFun}
    \begin{split}
        O(\bm{w_0}, &\cdots, \bm{w_{K}}) = \sum_{k=0}^{K} L(T_l^k; \bm{w_k}) + \\ &\sum_{k=1}^{K} \lambda_{k}  D(pred(T_u^{0}; \bm{w}_{0}), pred(T_u^{k}; \bm{w}_{k}))
    \end{split}
\end{equation}
The first term in the objective function is the loss over labeled samples at each resolution (fine and auxiliary coarse), which could take a standard form like cross-entropy loss. The second term in the loss function is a regularization term that enforces the consistency of predictions between the fine resolution and each of the coarse resolutions on the unlabeled instances i.e.; $L(T_l^k; \bm{w_k})$ is the loss over the labeled training instances $T_l^k$ on the $k$th resolution and the function $D()$ captures the consistency of the predictions between a pair of resolutions (fine, coarse) on unlabeled data. $\lambda_k$ for each coarse resolution $k$ is a hyperparameter fixed using cross-validation. The optimal parameters $\bm{w_0}, \cdots, \bm{w_{K}}$ are learned using gradient descent. Next, we design the consistency function $D$ that captures the many-to-one correspondence between instances of different resolutions.

\subsection{Defining consistency across resolutions}
Defining the consistency function $D$ is non-trivial because of the lack of an one-to-one mapping between instances across a pair of resolutions, as would be the case in traditional multi-view problems. Given a pair of resolutions - coarse and fine, one can define a many-to-one mapping of instances from the fine resolution to the coarse resolution by using a nearest neighbor approach. i.e.; every fine resolution instance is assigned to the coarse resolution instance with the closest location to it. Subsequently, the consistency of predictions on the unlabeled instances between a pair of resolutions boils down to defining the consistency between every coarse resolution instance and its corresponding fine resolution instances. In particular, the consistency term in equation \ref{eqObjFun} can be rewritten as, 
\begin{equation*}
\begin{split}
    D(&pred(T_u^{0}; \bm{w}_{0}), pred(T_u^{k}; \bm{w}_{k})) = \\
    &\sum_{i \in T_u^{k}}  d(\bm{x_i^{k}}, \{\bm{x_j^{0}} | j \in S_i \}, \bm{w_{k}}, \bm{w_{0}})
\end{split}
\end{equation*}

where the summation is over all unlabeled instances $i$ in the coarser resolution $R_k$. Also, the set $S_i$ denotes the unlabeled instances in fine resolution $R_0$ that are closest in location to instance $i$ from the coarser resolution. Thus, the objective function can be rewritten as,
\begin{equation}
    \label{eqObjFun2}
    \begin{split}
        &O(\bm{w_0}, \cdots, \bm{w_{K}}) = \sum_{k=0}^{K} L(T_l^k; \bm{w_k}) + \\ &\hspace{-2mm}\sum_{k=1}^{K} \hspace{-1mm}\lambda_{k} \hspace{-3mm} \sum_{i \in T_u^{k}}  d(\bm{x_i^{k}}, \{\bm{x_j^{0}} | j \in S_i \}, \bm{w_{k}}, \bm{w_{0}})
    \end{split}
\end{equation}
In this work, we propose two strategies to define the consistency of predictions $d()$ between a coarse resolution instance and its corresponding fine resolution instances: (1) Multi-instance learning and; (2) Attention mechanism

\subsubsection{Multiple Instance Learning (MIL) solution}
\begin{figure}
    \centering
    \includegraphics[width=1\linewidth]{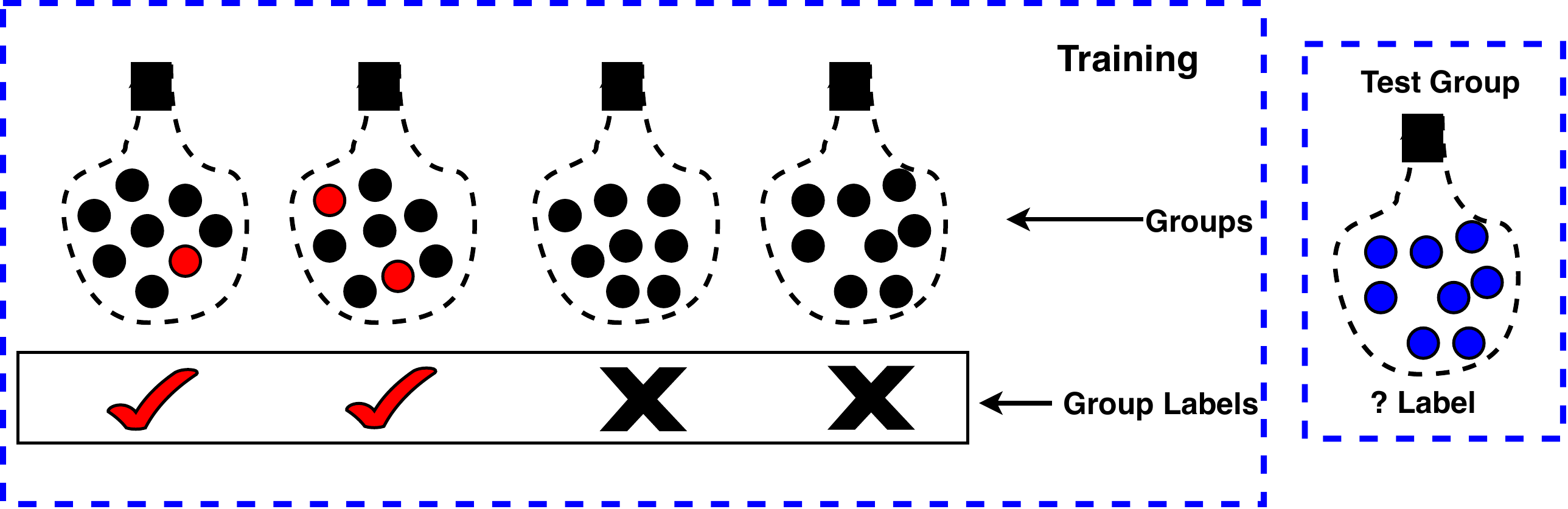}
    \caption{Presence-based assumption in Multiple Instance Learning (MIL) settings. If a group has a positive label, at least one instance within it is positive. Conversely, a negative group is made of all negative instances.}
    \label{fig:MILsetting}
\end{figure}

Multiple Instance Learning (MIL) \cite{MILSurvey1, MILSurvey2} considers the problem of learning predictive models to label groups of instances, in contrast to traditional settings where the goal is to label individual instances. Typical MIL classification settings operate under the presence-based assumption\cite{MILSurvey1} which states that a group has a positive label when at least one of its constituting instances has a positive label and it has a negative label when all of its constituting instances have a negative label as well. Figure \ref{fig:MILsetting} shows a caricature for the MIL assumption. MIL forms an direct way to define function $d()$ in equation \ref{eqObjFun2} that models the many-to-one relationship between an instance of a coarse resolution and its corresponding instances in the finer resolution. Given an instance $i$ from a coarse resolution $k_1$ and its corresponding instances $S_i$ from a fine resolution $k_2$, the prediction for the coarse resolution label can be written in two ways - first using the model on resolution $k_1$

\begin{equation*}
    Pr(y_i^{k_1} | \bm{x_{i}^{k_1}}) = f_{k_1}(\bm{x_{i}^{k_1}}; \bm{w}_{k_1})
\end{equation*}
Secondly, the label at $k_1$ can also be predicted using corresponding instances on $k_2$ using the MIL assumption.
\begin{equation*}
    Pr(y_i^{k_1} | \{\bm{x_{j}^{k_2}} | j \in S_i\} ) = \max_{j \in S_i} f_{k_2}(\bm{x_{j}^{k_2}}; \bm{w}_{k_2})
\end{equation*}
Note that taking the maximum of the probabilities for  instance-level for constituting instances is one way to implement the presence-based MIL assumption\cite{MILR}. Thus, the function $d()$ in equation \ref{eqObjFun2} can be defined as,
\begin{equation*}
\begin{split}
    d(\bm{x_i^{k_1}}, &\{\bm{x_j^{k_2}} | j \in S_i \text{ and } j \in T_u^{k_2} \}, \bm{w_{k_1}}, \bm{w_{k_2}}) = \\
    &\displaystyle\left(f_{k_1}(\bm{x_{i}^{k_1}}; \bm{w}_{k_1}) - \max_{j \in S_i} f_{k_2}(\bm{x_{j}^{k_2}}; \bm{w}_{k_2})\displaystyle\right)^2
\end{split}
\end{equation*}
Since we use gradient descent algorithm to learn the optimal parameters for our models, we want our objective functions to be differentiable and hence, in our implementation, the max function is replaced by its differentiable softmax approximation.

\subsubsection{Attention mechanism solution}
\begin{figure*}
    \centering
    \includegraphics[width=0.9\linewidth]{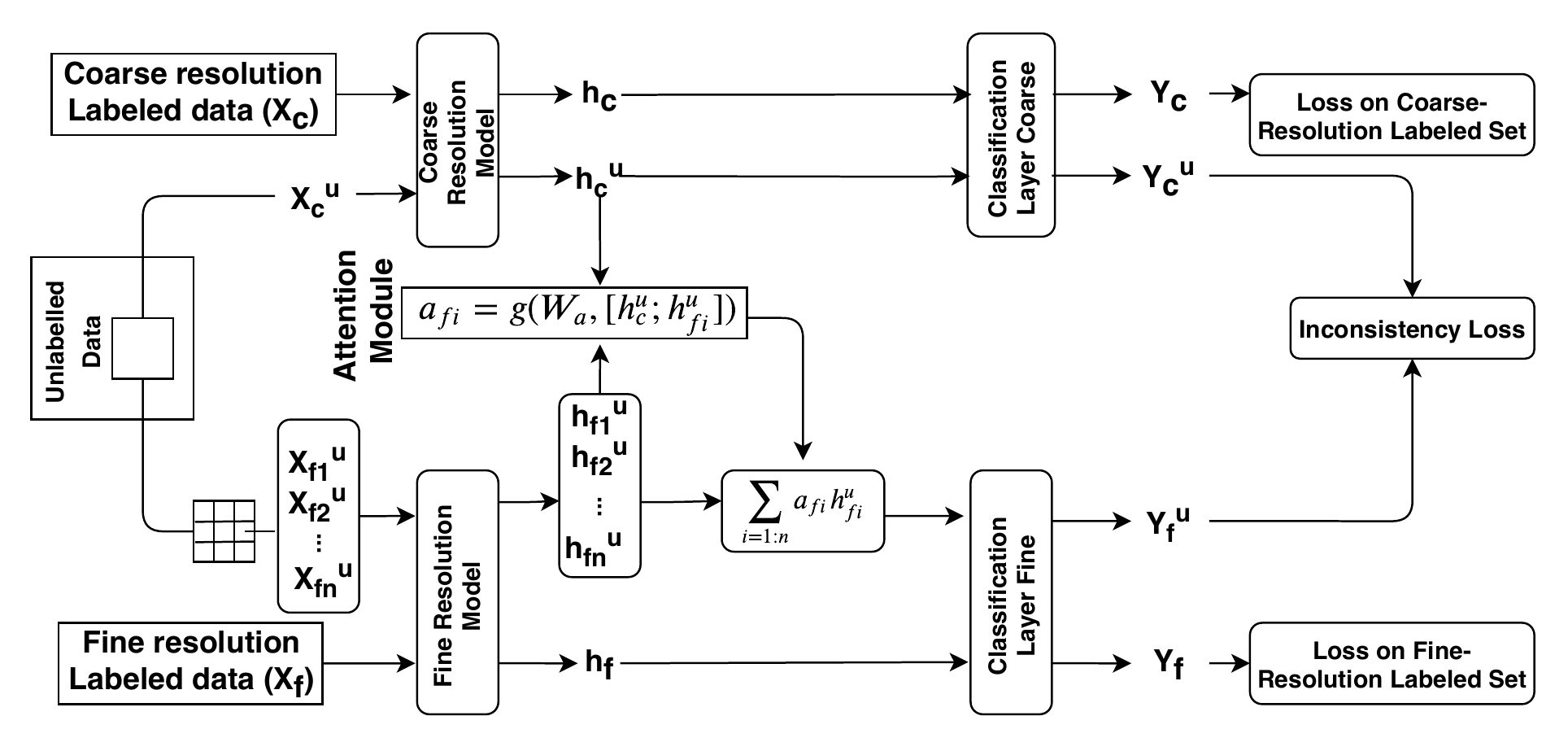}
    \caption{Attention mechanism based solution to the multi-resolution classification problem. Rather than assuming that a fixed number of instances at the finer resolution contribute to the label at the coarse resolution, this model adaptively predicts instances at the finer resolution that are best suited to predict this particular coarse resolution instance.}
    \label{fig:attentionMethods}
\end{figure*}
While the multiple-instance learning approach is a logical way to model the many-to-one relationship between instances of different resolutions, it might be a sub-optimal way to share information through this many-to-one correspondence. There could be a large variation in the number of positive fine resolution instances within a coarse resolution instance. Thus, by defining correspondence based on just one instance (with the maximum probability) or even a fixed fraction of instances does not capture the exact relationship between the two resolutions. Instead, we propose to parametrize the learning of this relationship so that we adaptively learn the most relevant instances on the finer resolution to predict the label on the coarser resolution. Specifically, we propose a solution to allow the model to attend to different fine resolution instances with different weights.
In our problem, we propose to use the concept of attention \cite{AttentionNLP} to assign relevance to fine resolution instances in the context of predicting the label on the corresponding coarser resolution instance. In particular, given a coarse resolution instance $i$ at resolution $k_1$ and corresponding fine resolution instances $S_i$ on resolution $k_2$, the attention weight $a_j$ for every instance $j \in S_i$ is defined as,
\begin{equation*}
    a_j^{k_2} = \frac{g([\bm{h_j^{k_2}}, \bm{h_i^{k_1}}]; \bm{w}_a)}{\sum_{k \in S_i}{g([\bm{h_k}, \bm{h_i}]; \bm{w}_{a{k_1}{k_2}})}}
\end{equation*}
where $h_j^{k_2}$ and $h_i^{k_1}$ are hidden representations for the instances obtained from models $f^{k_2}$ and $f^{k_1}$ respectively. For simpler models these could just be $\bm{x}_j^{k_2}$ and $\bm{x}_i^{k_1}$. Also, square brackets such as $[\bm{h_j^{k_2}}, \bm{h_i^{k_1}}]$ denotes vector concatenation of the elements within. Given these attention weights, a representation for all instances in $S_i$ can be computed as $\sum_{j \in S_i} a_j^{k_2} h_j^{k_2}$. This aggregated representative hidden state can then be fed into model $f^{k_2}$ to obtain the prediction for $Pr(y_i^{k_1} | \{\bm{x_{j}^{k_2}} | j \in S_i\} )$. The definition for consistency function $d()$ in equation \ref{eqObjFun2} is the similar as in the MIL case i.e.; 
\begin{equation*}
\begin{split}
    d(\bm{x_i^{k_1}}, &\{\bm{x_j^{k_2}} | j \in S_i \text{ and } j \in T_u^{k_2} \}, \bm{w_{k_1}}, \bm{w_{k_2}}, \bm{w}_{a{k_1}{k_2}}) \\
    =&\displaystyle\left(Pr(y_i^{k_1} | \bm{x_{i}^{k_1}}) - Pr(y_i^{k_1} | \{\bm{x_{j}^{k_2}} | j \in S_i\} )\displaystyle\right)^2 \\
    =& \displaystyle\left(f_{k_1}(\bm{x_{i}^{k_1}}; \bm{w}_{k_1}) - Pr(y_i^{k_1} | \sum_{j \in S_i} a_j^{k_2} h_j^{k_2} )\displaystyle\right)^2 \\
    =& \displaystyle\left(f_{k_1}(\bm{x_{i}^{k_1}}; \bm{w}_{k_1}) - f_{k_2}^{'}(\sum_{j \in S_i} a_j^{k_2} h_j^{k_2}; \bm{w}_{k_2})\displaystyle\right)^2
\end{split}
\end{equation*}
where $f_{k_2}^{'}$ is the part of the model $f^{k_2}$ that generates the predictions $y^{k_2}$ once the hidden states $h^{k_2}$ are computed. Figure \ref{fig:attentionMethods} shows a sketch of the proposed attention based model and the computation of the corresponding objective function.

\section{Evaluation}
We evaluate the performance of the proposed algorithms on 4 real world data sets. Three of these data sets come from the application of urban area mapping in the remote sensing domain. The fourth data set considers the task of sentiment classification in the natural language processing domain. Table \ref{tableDataSetSummary} shows the characteristics of these data sets. We only consider one coarse resolution in all of our data sets in this section, however, our methods are easily applicable when there is more than one source of coarse resolution data.
\begin{table*}[t]
    \centering
    \caption{Summary of data sets used in this paper}
        \begin{tabular}{|p{1.2cm}|p{5cm}|c|c|c|c|c|}
            \hline
            \multirow{2}{*}{\textbf{Dataset}} & \multirow{2}{*}{\textbf{Task}} & \multicolumn{2}{c|}{\textbf{Coarse}} & \multicolumn{3}{c|}{\textbf{Fine}}\\\cline{3-7}
            & &\textbf{Train} & \textbf{Unlabeled} & \textbf{Train} & \textbf{Unlabeled} & \textbf{Test}\\\hline\hline
            D1 & Urban mapping, Minneapolis& 2000 & 10000 & 200 & 90000 & 60000\\\hline
            D2 & Urban mapping, Madrid& 2000 & 10000 & 200 & 90000 & 60000\\\hline
            D3 & Urban mapping, Rome& 2000 & 10000 & 200 & 90000 & 60000\\\hline
            D4 & Sentiment classification, IMDB reviews& 2000 & 1000 & 200 & 10000 & 236\\\hline
        \end{tabular}
    \label{tableDataSetSummary}
\end{table*}
\setlist{nolistsep}
\begin{enumerate}
    \item \textbf{Urban mapping:} Data sets D1, D2, D3 in table \ref{tableDataSetSummary} correspond to the problem of urban mapping using satellite data for the cities Minneapolis, Madrid and Rome respectively. Our task in these problems is to use satellite-collected observations at a given location to determine if that location is urban or not. Different satellites observe the Earth at different spatial resolutions. In each one of our data sets, we consider observations from two publicly available \cite{lpdaac} satellite sources - Landsat and Sentinel. Landsat data is coarser (30m $\times$ 30m pixels) and has more labeled data while Sentinel data is finer (10m $\times$ 10m pixels) and has less supervision. Our goal is to detect urban areas at Sentinel resolution. Training labels for this task are handcrafted using data obtained from Open Street Maps (OSM).
    \item \textbf{Sentiment classification:} Data set D4 in table \ref{tableDataSetSummary} corresponds to the sentiment classification task on movie reviews. Our goal in this problem is to determine the sentiment (positive or negative) at fine resolution i.e at the sentence-level using auxiliary data at the coarse-resolution i.e at the review-level. We use the Large Movie Review Dataset created by the authors in \cite{imdb}. This data set has sentiment labels at the review-level. For sentence-level supervision, we use the limited labels curated by the authors in \cite{imdbSentence} for the same review corpus.
\end{enumerate}
All data sets will be made available through the author's personal website after publication, along with the code for the algorithms proposed in this paper.

\subsection{Comparison with baselines}
For our experiments, we compare the performance of our methods (\emph{Multi-Res MIL} and \emph{Multi-Res Attention}) against the following baselines -
\begin{enumerate}
    \item \textbf{OnlyFine} This method considers only labels from the fine resolution to train the model 
    \item \textbf{SSRManifold \cite{SSRManifold}} This method uses labeled and unlabeled data from the fine resolution. No coarse resolution data is used. Manifold regularization is a class of semi-supervised techniques where unlabeled instances close to one another in the feature space are expected to have similar predictions. Here, we use the formulation proposed in \cite{SSRManifold} using Euclidean distance and assuming a fully connected graph for computing neighborhood.
    \item \textbf{Propagate} This method uses coarse-resolution predictions to create pseudo-labels for the fine resolution. First, a classification model is learned on the coarse resolution and predictions are generated on the abundant unlabeled data on the coarse resolution. Now, these predictions are propagated onto the finer resolution to create pseudo-labels. If a coarse resolution instance is predicted positive, all fine resolution instances that correspond to it get a positive pseudo-label (and vice-versa).
    \item \textbf{Augment \cite{augment}} This method, proposed in \cite{augment}, considers the problem of classification on multiple resolutions. In this method, the training set at each resolution is iteratively grown to include unlabeled instances that have consistent predictions between resolutions. Models are retrained after each iteration with the modified training set. In this case, consistency between resolutions is defined using the presence-based assumption of Multiple Instance Learning (MIL).
\end{enumerate}
\subsubsection{Base model}
The ideas presented in this paper can be used with any base model. To keep comparisons fair, we use the same base classifier for all baselines and our two proposed methods.
\begin{enumerate}
    \item \textbf{Urban mapping: } Here, we use a neural network with one hidden layer as the base model at each resolution.
    \item \textbf{Sentiment classification: } Here, we use LSTM (Long short-term memory) \cite{lstm} as the base model at each resolution. Individual words are encoded using \emph{word2vec} embeddings \cite{word2vec} before feeding it into the LSTM model.
\end{enumerate}
\begin{table}[h]
  \centering
  \caption{Comparison with baselines: the average accuracy and standard deviation over 5 runs is reported for different algorithms }
  \begin{tabular}{|c|c|c|c|c|}
    \hline
    \textbf{Method} & \textbf{D1} & \textbf{D2} & \textbf{D3} & \textbf{D4}\\\hline\hline
    OnlyFine & $\underset{(0.111)}{0.59}$ & $\underset{(0.059)}{0.68}$ & $\underset{(0.066)}{0.64}$ & $\underset{(0.011)}{0.62}$\\ \hline
    SSRManifold & $\underset{(0.079)}{0.63}$ & $\underset{(0.041)}{0.70}$ & $\underset{(0.049)}{0.69}$ & $\underset{(0.015)}{0.64}$\\ \hline
    Propagate & $\underset{(0.082)}{0.71}$ & $\underset{(0.044)}{0.81}$ & $\underset{(0.008)}{0.82}$ & $\underset{(0.012)}{0.66}$\\ \hline
    Augment & $\underset{(0.117)}{0.69}$ & $\underset{(0.072)}{0.72}$&  $\underset{(0.042)}{0.73}$ & $\underset{(0.015)}{0.66}$\\ \hline
    \begin{tabular}{@{}c@{}}Multi-Res \\ MIL\end{tabular} & $\underset{(0.087)}{0.69}$ & $\underset{(0.096)}{0.69}$ & $\underset{(0.012)}{0.66}$ & $\underset{(0.066)}{0.64}$\\ \hline
    \begin{tabular}{@{}c@{}}Multi-Res \\ Attention\end{tabular} & $\underset{(0.0068)}{\textbf{0.85}}$ & $\underset{(0.025)}{\textbf{0.90}}$ & $\underset{(0.022)}{\textbf{0.91}}$ & $\underset{(0.013)}{\textbf{0.70}}$\\ \hline
\end{tabular} 
    \label{tableResults}
\end{table}
\subsubsection{Results}
Table \ref{tableResults} reports the average accuracy and standard deviation over 5 iterations for different algorithms on the data sets in table \ref{tableDataSetSummary}. As expected, \emph{OnlyFine} does not perform well since it only uses the limited supervision available on the fine resolution. \emph{SSRManifold} goes one step ahead and uses unlabeled data in addition to labeled data from the fine resolution. Thus, its performance is slightly better than \emph{OnlyFine}. Next, we have methods \emph{Propagate} and \emph{Augment} that try to sub-optimally combine the data from two resolutions. \emph{Propagate} uses the pseudo-labels generated from the coarse-resolution model predictions for training the fine resolution model. Since not every fine-resolution instance within a positive coarse-resolution instance is positive (and vice-versa), these pseudo-labels are inaccurate and treating them as ground-truth to train the fine resolution model is sub-optimal. \emph{Augment} also tries to leverage information from other resolutions, but it uses a max-aggregation from MIL to model the many-to-one correspondence between resolutions to check for confident instances in the unlabeled data. This is clearly not optimal as the number of corresponding instances in the fine resolution of the same label as the coarse resolution instance can vary. Moreover, this greedy approach of simply adding the instances where the models on different resolutions agree, to the labeled set is inferior compared to continuously updating the models based on consistency on unlabeled data, that the \emph{Multi-Res} models proposed in this paper do. \emph{Multi-Res MIL} uses the MIL assumption as well, so it performs similar to \emph{Augment}. However, \emph{Multi-Res Attention} uses the feature values to determine most relevant instances to assign correspondence between instances of different resolutions, thus making it very flexible to the number of positive fine-resolution instances within a positive coarse-resolution instance (and vice-versa). This is reflected in its performance gain over other baseline methods.

\subsection{Explaining predictions through attention weights}
\begin{figure}[h]
    \begin{subfigure}{0.45\linewidth}
        \centering
        \includegraphics[width=0.9\linewidth]{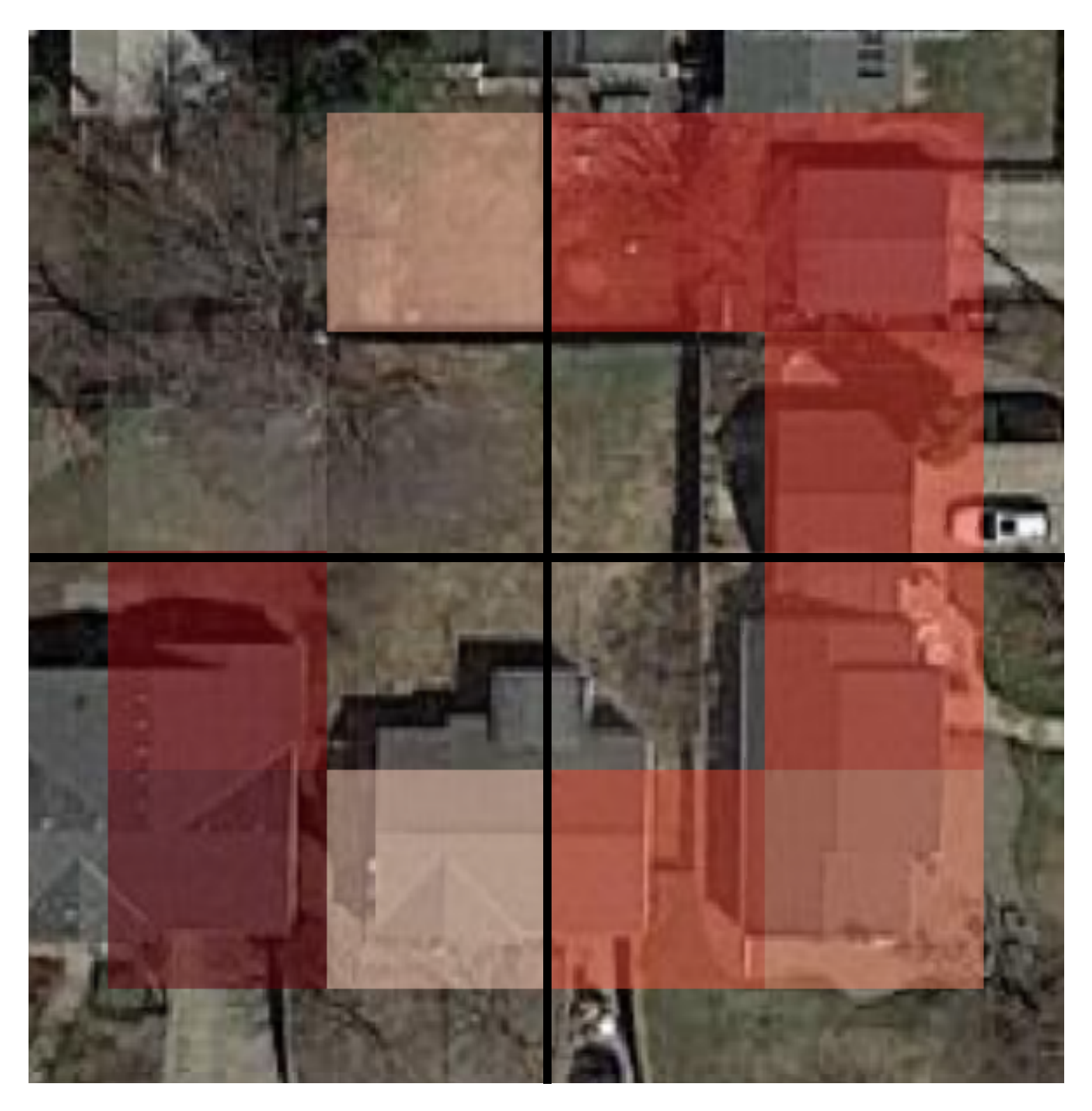}
        \caption{}
        \label{urbanMapping_Att1}
    \end{subfigure}
    ~
    \begin{subfigure}{0.45\linewidth}
        \centering
        \includegraphics[width=0.9\linewidth]{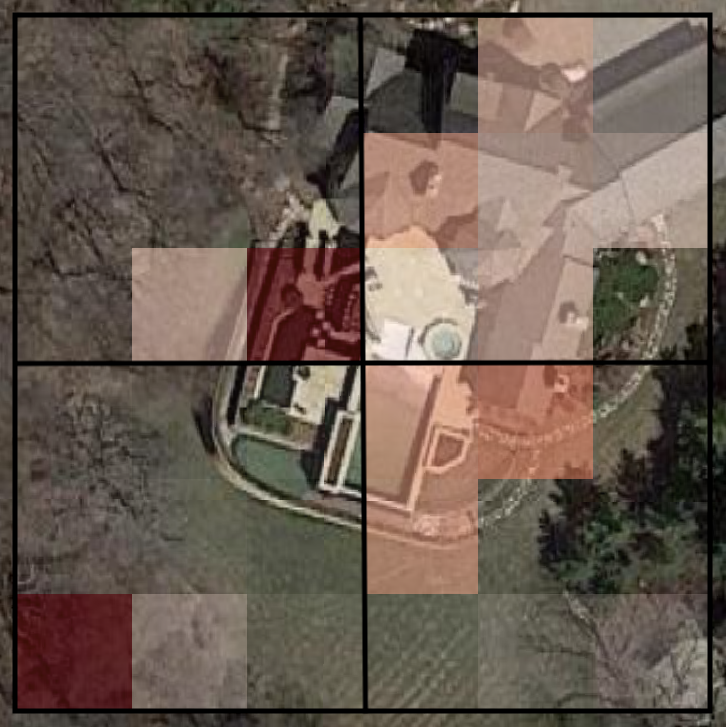}
        \caption{}
        \label{urbanMapping_Att2}
    \end{subfigure}
    
    \begin{subfigure}{\linewidth}
        \centering
        \includegraphics[width=\linewidth, height = 67mm]{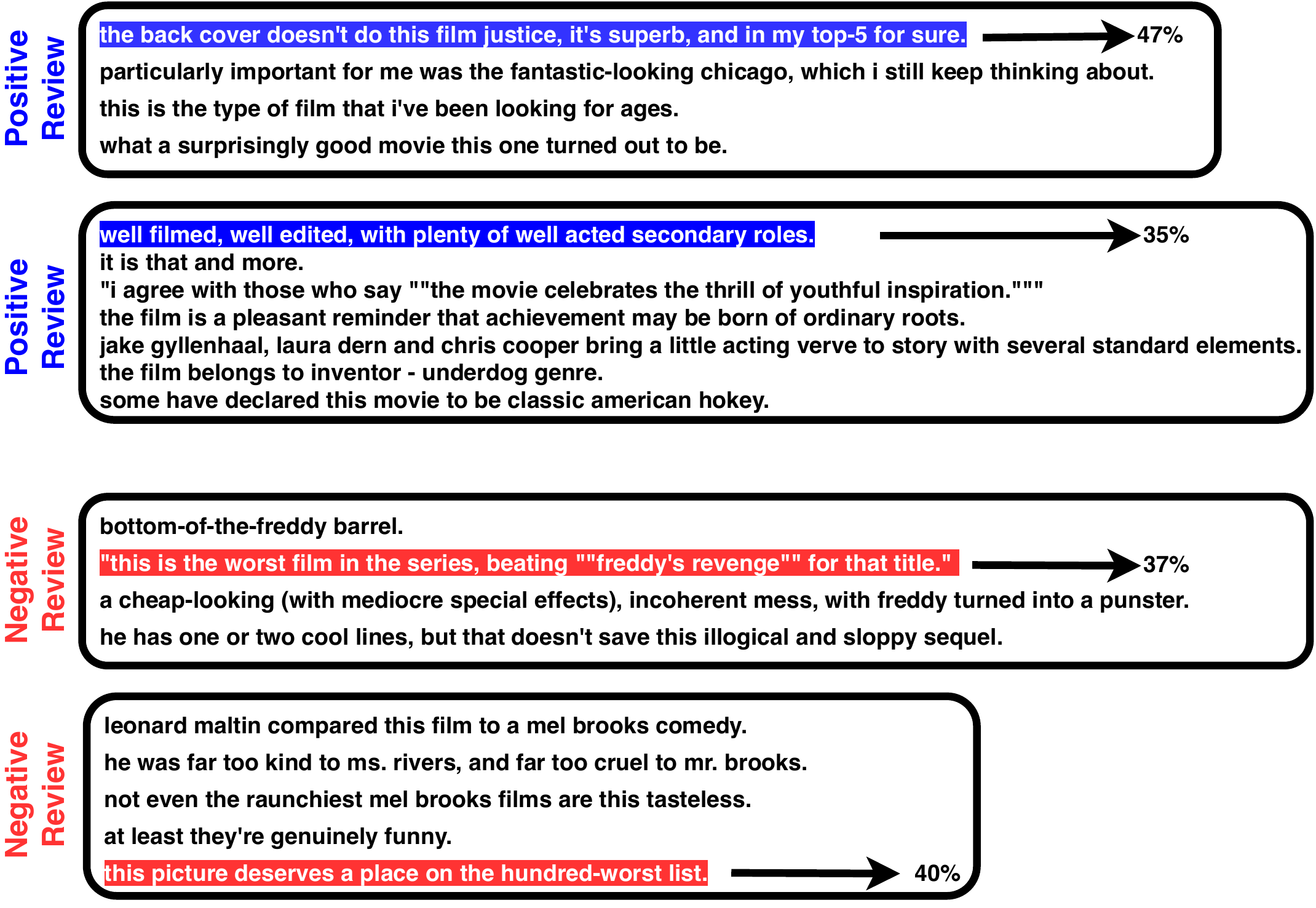}
        \caption{}
        \label{sentimentClassification_Att}
    \end{subfigure}
    \caption{Attention weights learned by the \emph{Multi-res Attention} model on examples from the urban mapping and sentiment classification applications. Attention weights are normalized so that the sum of weights on fine resolution instances corresponding to a coarse-resolution instance sum to 1. In the case of urban mapping, there are 9 fine resolution instances within a coarse resolution instance. Boundary of each coarse resolution pixel is highlighted in black. The more red a fine resolution instance looks, the more weight it has. In the sentiment classification examples, the sentence with the most attention is highlighted. Colors blue and red denote positive and negative sentiment respectively.}
    \label{fig:Att_Weights_Examples}
\end{figure}
One of the key challenges in applying advanced machine learning models in many real-world applications is their lack of interpretability. . Since attention mechanism computes relevance over different elements, it has been found useful to identify the key elements that led to a given prediction from the model \cite{attentionMIL}, thereby increasing its explainability. Since our approach to use coarse-resolution data to improve fine-resolution models also uses attention mechanism, we can expect to see some explainability and confidence added to the model predictions by examining the corresponding attention weights. Figure \ref{fig:Att_Weights_Examples} shows the attention weights learned for a few examples from both applications: urban mapping and sentiment classification. In the urban mapping examples, there are 9 fine resolution pixels within each coarse resolution pixel. Both examples have 4 coarse resolution pixels, whose boundaries are marked black. The attention weights are normalized for instances within each coarse resolution pixel, so that they sum to 1. The darker shade of red the fine resolution pixel looks, the higher attention weight it has. Since attention weights are used to assign correspondence, we would expect that a urban coarse resolution pixel would have higher attention weights on urban fine resolution pixels and vice-versa. This can be clearly seen in the examples. In the  example in figure \ref{urbanMapping_Att2}, the house is divided in 4 coarse resolution pixels, 3 of which seem to have multiple fine resolution urban pixels, from a visual inspection of the underlying Google Earth imagery. Thus, these coarse-resolution pixels will be predicted urban. Within these pixels, we can see that all the attention gets distributed among the fine resolution pixels that lie within the area of the house. The fine resolution pixels outside the area of the house get no attention within these urban coarse-resolution pixels. However, the bottom left coarse resolution pixel in this example has only a tiny overlap with the house and hence is non-urban. We can see that in this case, the model chose to put all its attention on a non-urban fine resolution pixel.

Similarly, the attention weights learned on the sentiment classification task offer interesting explanations to the model predictions. The figure lists 2 positive and 2 negative reviews. The sentence that carried the most attention weight is highlighted. The examples clearly show the attention being put on the sentences that reflect the most synergy with the overall sentiment of the whole review. From the results in this section, we can conclude that, if coarse resolution data was also available for test instances, we can inspect the learned attention weights and use the consistency between resolutions to add confidence in the model predictions.

\subsection{Effect of increasing labeled samples}
\begin{figure}[h]
    \centering
    \includegraphics[width=\linewidth]{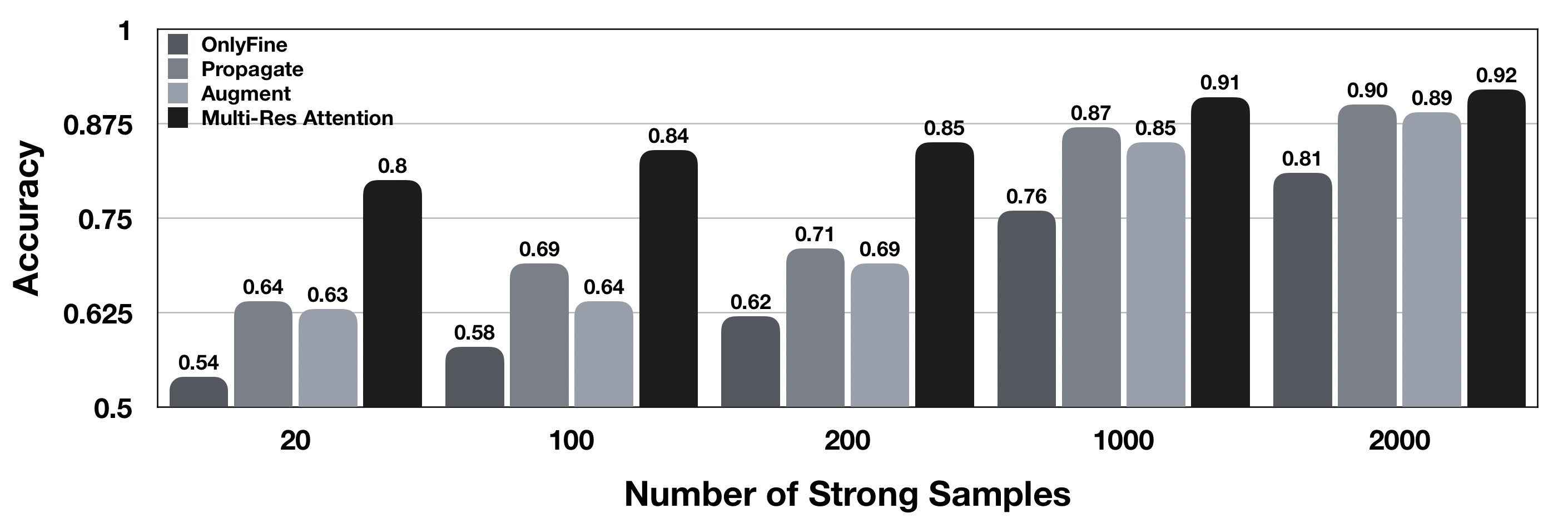}
    \caption{Gain with increasing number of fine-resolution labeled samples: The x-axis shows the number of labeled samples provided to the algorithms.}
    \label{fig:NumSamples}
    \vspace{-7mm}
\end{figure}

\emph{Multi-Res Attention} tries to leverage the labeled data from coarser resolutions along with abundant unlabeled data to improve the performance of a fine-resolution model that just uses limited labeled data on the fine resolution. As the number of labeled instances on the fine resolution increases, the gain obtained by using the auxiliary information from coarser resolutions reduces.  We demonstrate this empirically by increasing the number of fine-resolution labeled instances in data set D1 from $20$ up to $2000$. From the results in figure \ref{fig:NumSamples}, we observe that the accuracy values for all methods increase as the number of fine-resolution labeled samples are increased. The benefit of using coarse-resolution information (over \emph{OnlyFine}) through methods \emph{Augment} and \emph{Propagate} are lower in comparison to \emph{Multi-Res Attention} for every value of the number of fine-resolution labeled samples. Finally, given enough number of labeled samples, \emph{OnlyFine} will catch up to \emph{Multi-Res Attention} in terms of its accuracy value.

\subsection{Increasing model complexity through semi-supervised learning}
\begin{figure}
\centering
\includegraphics[width=0.4\textwidth]{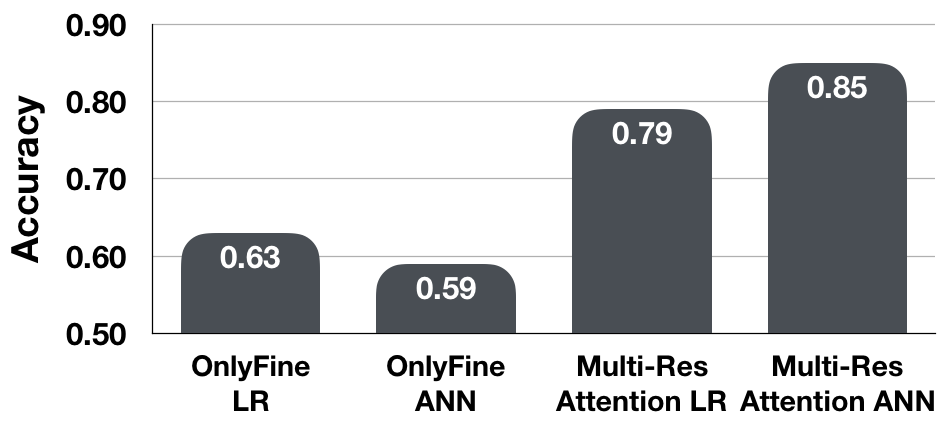}
\caption{Performance of a standard logistic regression (LR) and a single- layer Artificial Neural Network (ANN) that uses only fine resolution labels as well as their counterparts that also make use of labels at multiple resolutions for Data set D1.}
\label{figModelComplexity}
\vspace{-5mm}
\end{figure}

A key motivation for using semi-supervised learning is that for many applications, simpler models do not suffice and we would like to use models with more parameters like deep neural networks. However, the more complex the model, the more samples needed to train it. Semi-supervised learning through auxiliary coarse-resolution information can be used to improve generalization performance of complex models that would otherwise have poor generalization performance due to limited fine-resolution labeled samples. To demonstrate this, we train a logistic regression model and an artificial neural network with one hidden layers on the data set D1. 
The ANN is a more complex model, so it is expected to overfit on the training data if we use only fine-resolution labeled samples. However, once we use labeled and unlabeled samples from the coarse resolution as well, it outperforms the logistic regression model, as shown in figure \ref{figModelComplexity}. 

\section{Conclusion}
In this paper, we formalized the semi-supervised learning problem of using labeled and unlabeled data from coarser resolutions to aid models trained to predict phenomena on fine resolution as a multi-view framework. The multi-view framework helps to regularize the models trained on individual resolutions by enforcing consistency of predictions across resolutions on the large number of freely-available unlabeled data. Unlike traditional multi-view learning scenarios, the multi-resolution classification task involves a many-to-one correspondence between views of the data, which the proposed methods in the paper learn explicitly through multiple instance learning and attention mechanism. Experiments on publicly-available remote sensing and natural language processing data sets show the utility of utilizing coarse-resolution data through the multi-view framework. The attention mechanism solution also offers interpretability that might be useful in adding confidence to model predictions. Future work involves integrating this attention-based multi-view approach with state-of-the-art classification models specific to sequence data such as in natural language processing or raster data such as in computer vision.

\end{document}